# Complexity Analysis of Vario-eta through Structure

Alejandro Chinea and Elka Korutcheva

Departamento de Física Fundamental, Facultad de Ciencias UNED,
Paseo Senda del Rey nº9, 28040-Madrid - Spain

**Abstract.** Graph-based representations of images have recently acquired an important role for classification purposes within the context of machine learning approaches. The underlying idea is to consider that relevant information of an image is implicitly encoded into the relationships between more basic entities that compose by themselves the whole image. The classification problem is then reformulated in terms of an optimization problem usually solved by a gradient-based search procedure. Vario-eta through structure is an approximate second order stochastic optimization technique that achieves a good trade-off between speed of convergence and the computational effort required. However, the robustness of this technique for large scale problems has not been yet assessed. In this paper we firstly provide a theoretical justification of the assumptions made by this optimization procedure. Secondly, a complexity analysis of the algorithm is performed to prove its suitability for large scale learning problems.

## 1 Introduction

In general, an image can always be interpreted as a combination or mixture of simpler entities. The complexity of image analysis is usually due to the difficulties involved in discovering the nature of the relationship of these constituent elements. Surprisingly, species ranging from insects to mammals solve visual recognition problems extremely well. They have an inherent capacity to represent the temporal structure of experience, such information representation making it possible for an animal to adapt its behaviour to the temporal structure of an event in its behavioural space [9]. In particular, humans can recognize an object from a single presentation of its image without the necessity of integrating information over multiple time steps as would be required by classic machine learning paradigms. Indeed, the human perception system organizes information by using cohesive structures [13, 14]. Specifically, the human perception process always tries to assign structure to any perceptual information based upon previously stored knowledge structures (e.g. an image is recognized by identifying its structural components). This fact validates a well-known premise from the machine learning field that states that the structure of an entity is very important for both classification and description. Furthermore, these ideas have motivated the development of a new branch of machine learning algorithms [7,8,10,15,17,19] that use structured representations (i.e. graph-based representations) of data for both classification and regression tasks instead of classic vector-based representations. For instance, an image can be represented by its region adjacency graph [1, 16] (see figure 1) which is extracted from the image by associating a node to each homogeneous region, and linking the nodes related to adjacent regions. In the region adjacency representation each node is labelled by a real vector, that represents features of the region (position, area, mean colour, texture, etc.). Thus, not only perceptual features of the image are captured with this representation but also its spatial arrangement. It is important to note that the notion of information content is strongly linked to the notion of structure.

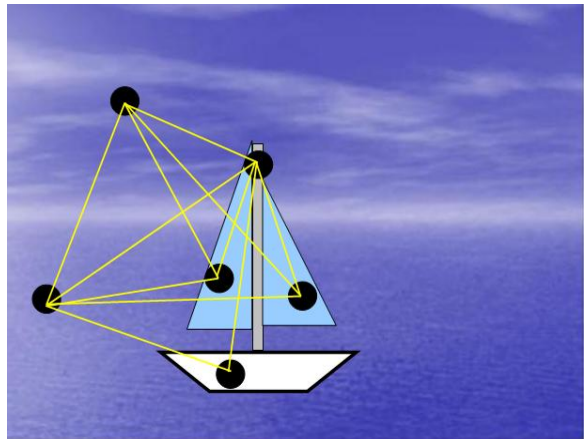

**Fig. 1.** Example of a graph-based representation of an artificial image using a region adjacency graph, each node of the graph is related to a homogeneous region of the image.

Moreover, image processing is done on the numerical representation of the image, i.e. a large matrix, therefore, these arrays can be enormous, and as soon as one deals with a sequence of images, such as in television or video applications, the volume of numerical data that must be processed becomes immense. Taking into account these considerations it can be deduced that the classification of large image data sets is a difficult computational task that becomes even more complex when dealing with structured representations of images. Therefore, from a machine learning point of view a great deal of research has been devoted to the development of fast computation learning schemes. In particular, for structured representations of data vario-eta through structure [3] has been recently proposed as an efficient learning scheme offering a good trade-off between speed of convergence and associated computational cost. Furthermore, this learning scheme achieves a rate of convergence similar to a quasi-Newton method at the computational cost of a first order method, and additionally, it offers the possibility of working in both sequential and in batch modes.

The present paper explores the reliability of this approximate second order stochastic gradient-based search optimization procedure for large scale learning problems. The rest of the paper is organized as follows: in the next section, some background topics on the vario-eta gradient-based optimization technique are introduced. Furthermore, the assumptions made by this optimization procedure are assessed from a theoretical point of view in order to check their reliability. Section 3 is devoted to the complexity analysis of the algorithm for structured domains. Specifically, the generating functions of the recursions associated to this learning scheme are analysed from an analytic combinatorics point of view. Section 4, is focused on the practical applications of the complexity analysis performed in section 3. Finally, section 5 provides a summary of the present study together with some concluding remarks.

## 2  Vario-eta Learning Rule

The problem of learning in both deterministic and probabilistic machine learning models is frequently formulated in terms of parameter optimization techniques. The optimization problem is usually expressed in terms of the minimization of an error function $E$. This error is a function of the adaptive parameters of the model. One of the simplest techniques is gradient-search optimization which is one which has been widely studied. Here we investigate an approximate second order stochastic gradient-search learning rule known as vario-eta [20].

### 2.1  Mathematical Description

Without a loss of generality, let us denote as $E(w)$ the error function to be minimized. In addition, let us suppose that we express the parameters of our model in a vector $w=[w_1,w_2,w_3,....,w_m]$. A perturbation of the error function around some point of the model parameters can be written as follows: $E(w+\Delta w) = E(w_1+ \Delta w_1, w_2+ \Delta w_2,..., w_m+ \Delta w_m)$. Considering the Taylor expansion of the error function around the perturbation $\Delta w$ we obtain:

$$E(w+\Delta w) = E(w) + \sum_{i=1}^{m} \frac{\partial E}{\partial w_i}(\Delta w_i) + \frac{1}{2!}\left(\sum_{i=1}^{m} \frac{\partial E}{\partial w_i}(\Delta w_i)\right)^2 +$$

$$+ \frac{1}{3!}\left(\sum_{i=1}^{m} \frac{\partial E}{\partial w_i}(\Delta w_i)\right)^3 + ..........$$

(1)

In the batch version of gradient-descent approach, we start with some initial guess of model parameters. Then, model parameters are updated iteratively using the entire data set. In the sequential [14, 15] version of gradient descent, the error function gradient is evaluated for just one pattern (or a batch of patterns smaller when compared to the size of the data set for the case of the vario-eta learning rule) at a time. Each update of model parameters can be viewed as perturbations (i.e. noise) around the current point given by the *m* dimensional vector of model parameters. Let us assume a given sequence of *N* disturbance vectors Δ*w*. Considering the error as a random variable and ignoring third and higher order terms in expression (1), the expectation of the error <*E(w)*> can be expressed as:

$$\langle E(w) \rangle \cong \frac{1}{N} \sum_{n=1}^{N} E(w + \Delta w^n) \qquad (2)$$

Substituting the Taylor expansion of the error function in expression (2) and rearranging terms we obtain a series expansion of the expectation of the error as a function of the moments of the random perturbations:

$$\langle E(w) \rangle \cong E(w) + \sum_{i=1}^{m} \langle \Delta w_i \rangle \frac{\partial E}{\partial w_i} + \frac{1}{2} \sum_{i=1}^{m} \langle (\Delta w_i)^2 \rangle \frac{\partial^2 E}{\partial w_i^2} + \sum_{i<j} \langle \Delta w_i \Delta w_j \rangle \frac{\partial^2 E}{\partial w_i w_j} \qquad (3)$$

In addition, the weight increment associated to the gradient descent rule is $\Delta w_i = -\eta g_i$. The third term of expression (3) concerning the covariance can be ignored supposing that the elements of the disturbance vectors are uncorrelated over the index n. This is a plausible hypothesis given that patterns of the data set are selected randomly during the optimization procedure. Moreover, close to a local minimum we can assume that $\langle \Delta w_i \rangle \cong 0$. Taking into account these considerations the expectation of the error is then given by:

$$\langle E(w) \rangle \cong E(w) + \frac{1}{2} \sum_{i=1}^{m} \langle (\Delta w_i)^2 \rangle \frac{\partial^2 E}{\partial w_i^2} = E(w) + \frac{\eta^2}{2} \sum_{i=1}^{m} \sigma^2(g_i) \frac{\partial^2 E}{\partial w_i^2} \qquad (4)$$

From equation (4), it is easy to deduce that the expected value of the error increases as the variance (represented by the symbol $\sigma^2$) of the gradients increases. This observation suggests that the error function should be strongly penalized around such model parameters. Therefore, to cancel the noise term in the expected value of the error function the gradient descent rule must be changed to $\Delta w_i = -\eta g_i/\sigma(g_i)$. This normalization is known as vario-eta and was proposed in [20] for training neural networks. Specifically, the learning rate is renormalized by the stochasticity of the error gradient signals.

## 2.2 Virtues and Limitations of the Approximation

Expression (4) was obtained under the hypothesis that the expectation of the perturbations $\Delta w_i$ could be neglected near to a local minimum. To confirm the validity of this approximation, let us firstly express the expectation of the perturbations $\Delta w_i$ in terms of the error gradients $g_i$ (using the expression of the gradient optimization update rule):

$$\langle \Delta w_i \rangle = \left\langle -\frac{\eta g_i}{\sigma(g_i)} \right\rangle = -\eta \frac{\langle g_i \rangle}{\sqrt{\langle g_i^2 \rangle - \langle g_i \rangle \langle g_i \rangle}} \tag{5}$$

$$\langle g_i \rangle = \frac{1}{N} \sum_{n=1}^{N} g_i^n \quad \langle (g_i)^2 \rangle = \frac{1}{N} \sum_{n=1}^{N} (g_i^n)^2 \tag{6}$$

Substituting expression (6) (i.e. the values of the first and second order moments of the error gradients) in expression (5) and rearranging, we get:

$$\langle \Delta w_i \rangle = \frac{-\eta}{\sqrt{N \frac{\sum_{n=1}^{N}(g_i^n)^2}{\left(\sum_{n=1}^{N} g_i^n\right)^2} - 1}} = \frac{-\eta}{\sqrt{\frac{N}{f(g_i^1, g_i^2, g_i^3, ...., g_i^N)} - 1}} \tag{7}$$

In the previous expression, we introduced the multidimensional function $f$. This function depends on the error gradients obtained through $N$ iteration steps. Furthermore, it is easy to show (see expression (7)) that the minimum and maximum values of the function are 1 and $N$ respectively. Specifically, the function reaches its maximum when the error gradients are identical.

$$f(g_i^1, g_i^2, g_i^3, ...., g_i^N) = \frac{\left(\sum_{n=1}^{N} g_i^n\right)^2}{\sum_{n=1}^{N} (g_i^n)^2} = 1 + 2 \frac{\sum_{n<k} g_i^n g_i^k}{\sum_{n=1}^{N} (g_i^n)^2} \tag{8}$$

This fact is of particular interest as it implies that the error gradients must be identical through $N$ iteration steps. Taking into account that $N$ represents the size of the batch, the probability of such event is $M^{-N}$, where $M$ is the total size of the training set. It is important to note that in the batch version of gradient descent $M=N$.

Therefore, such event can be ignored as its probability can be considered zero as in practice $M>>N>>1$. However, numerical precision errors could make the contribution of the term (7) big enough to break the validity of the hypothesis. This scenario would cause oscillations of the error $E$ during the optimization procedure. Therefore, the approximation is limited in practice by the precision of the machine; although their impact is minimal, taking into account the precision achieved by current 64 bit platforms of modern computers. Moreover, in order to alleviate eventual numerical problems a small constant $0 < \phi << 1$ can be summed to the standard deviation of the error gradients (e.g. $\phi = 10^{-6}$) as suggested in [3].

## 3 Complexity Analysis

From a computational point of view machine learning models that work with structured representations of data are intrinsically more complex than their vector-based counterparts are. The fact of using structured representations of data is translated into a substantial gain in information content but also in an increase on computational complexity. Indeed, the principal drawback of these models is an excessively expensive computational learning phase. The learning rule studied in the previous section was adapted and expressed in a recursive form in [3] for working with structured representations of data. The fact of expressing such calculations in a recursive form permitted a considerable reduction in memory storage something fundamental for learning problems composed of huge data sets.

In this section, a complexity analysis of such algorithm is performed using elements of the theory of analytic combinatorics [4, 6]. The main objective of analytic combinatorics is to provide quantitative predictions of the properties of large combinatorial structures. This theory has emerged over recent decades as an essential tool for the analysis of algorithms and for the study of scientific models in many disciplines.

Here, our intention is to study the asymptotic behaviour of the learning rule to reduce its computational requirements for large-scale learning problems. To this end, in subsection 3.1, we firstly provide some background topics on analytic combinatorics. Afterwards, in subsection 3.2 we model the algorithmic structure of the learning rule in terms of generating functions. Finally, in subsection 3.3 the singularities of the generating functions are analysed aimed at obtaining an asymptotic approximation of its coefficients for reducing the computational requirements associated to this learning rule.

### 3.1 Theoretical Background

Let us introduce some basic definitions to be used throughout the rest of this paper:

Definition 3.1.1: The ordinary generating function of a sequence $A_n$ is the formal power series:

$$A(z) = \sum_{n \geq 0}^{\infty} A_n z^n \qquad (9)$$

Generating functions are the central object of study of analytic combinatorics. Their algebraic structure directly reflects the structure of the underlying combinatorial object. Furthermore, they can be viewed as analytic transformations in the complex plane and its singularities account for the asymptotic rate of growth of function's coefficients. In addition, the theory elaborates a collection of methods (e.g. singularity analysis or saddle point method) by which one can extract asymptotic counting information from generating functions.

Definition 3.1.2: we let generally $[z^n]f(z)$ denote the operation of extracting the coefficient of $z^n$ in the formal power series $f(z) = \sum f_n z^n$

$$[z^n]\left(\sum_{n \geq 0}^{\infty} f_n z^n\right) = f_n = \frac{1}{2\pi i} \oint_c \frac{f(z)}{z^{n+1}} dz \qquad (10)$$

In expression (10), Cauchy's integral formula expresses coefficients of analytic functions as contour integrals. Therefore, an appropriate use of Cauchy's integral formula then makes it possible to estimate such coefficients by suitably selecting an appropriate contour of integration.

### 3.2 Vario-eta Generating Functions

Generally speaking, an algorithm can be interpreted as a mathematical object that is built iteratively from a set of finite rules that work on finite data structures. Furthermore, they have an inherent combinatorial structure that can be modelled in terms of generating functions. In the following, we model the algorithmic description of the vario-eta learning rule in terms of generating functions.

The algorithmic description of the learning rule provided in [3] is composed of two recurrences expressed in matrix form. These two finite differences equations accounted for the calculation of the mean and variance of the error gradients during the gradient-descent optimization loop. In order to simplify the notation, let us write without a loss of generality such equations in a single variable form:

$$\hat{g}_n = a_n \hat{g}_{n-1} + b_n g_n \qquad \Leftrightarrow \qquad \hat{g}_n = \frac{1}{n}\sum_{k=1}^{n} g_k \qquad (11)$$

$$\sigma_n^2 = a_{n-1}\sigma_{n-1}^2 + b_n(g_n - \hat{g}_{n-1})^2 \quad \Leftrightarrow \quad \sigma_n^2 = \frac{1}{n-1}\sum_{k=1}^{n}(g_k - \hat{g}_n)^2 \qquad (12)$$

The generating functions associated to the recurrences for the mean and variance of the error gradients can be computed applying the transformation (9) to both sides of equations (11) and (12). Specifically, equation (12) was expressed only in terms of the averaged error gradients and their variance to eliminate its dependency with the unknown generating function of the error gradients g(z).

$$\sum_{n\geq 0}^{\infty}\hat{g}_n z^n = \sum_{n\geq 0}^{\infty}(1-\frac{1}{n})\hat{g}_{n-1} z^n + \sum_{n\geq 0}^{\infty}\frac{1}{n} g_n z^n \tag{13}$$

$$\sum_{n\geq 0}^{\infty}\sigma_n^2 z^n = \sum_{n\geq 0}^{\infty}\sigma_{n-1}^2 z^n - \sum_{n\geq 0}^{\infty}\frac{1}{n-1}\sigma_{n-1}^2 z^n + \sum_{n\geq 0}^{\infty} n\hat{g}_n^2 z^n$$
$$-2\sum_{n\geq 0}^{\infty} n\hat{g}_n \hat{g}_{n-1} z^n \tag{14}$$

After doing some algebra and using the complex convolution theorem [18], we obtain the following integral equations in the complex variable z for the generating functions of mean of error gradients $\hat{g}(z)$ and their variance $\sigma^2(z)$ respectively:

$$\hat{g}(z) = \int \frac{1}{z(1-z)} g(z) dz \tag{15}$$

$$(1-z)\sigma^2(z) + z\int \frac{1}{z}\sigma^2(z) dz = \frac{1}{2\pi i}\oint_c (1-u)\frac{d\hat{g}(u)}{du}\hat{g}(z/u) du \tag{16}$$

Despite we do not know the form of the generating function of the error gradients neither their counting sequence $g_n$ we can make reasonable hypothesis about their form turning into a probabilistic framework. More specifically, the counting sequence $g_n$ is the result of computing the error gradient associated to a pattern randomly selected at iteration step *n* from the training set during the optimization loop. Therefore, we can consider each of them as *n* independent random variables. Furthermore, during the optimization loop the parameters of the machine learning model are not updated, therefore the *n* independent random variables representing the error gradients will be statistically equally distributed. It is important to note that the machine learning model can be viewed as a functional of the optimization parameters. Moreover, error gradients are calculated evaluating the functional by using the patterns of the data set. Taking into account we are interested in large-scale learning problems the statistical distribution of the averaged error gradients $\hat{g}_n$ will tend, by virtue of Laplace's central-limit theorem, to a Gaussian distribution at the extent the value of *n* increases. It is important to note that the value *n* represents the size of the data set for the case of batch learning and the size of the batch of patterns for sequential learning problems. Thus, its probability generating function will have the following expression:

$$\hat{g}(z) \cong e^{\mu z - \frac{1}{2}\sigma^2 z^2} \tag{17}$$

The complex function $\hat{g}(z)$ is analytic for $|z| < \infty$, it is also holomorphic (i.e. complex-differentiable), which is equivalent to saying that it is also analytic. Additionally, it can be proved that $\hat{g}(1/z)$ is also analytic. Moreover, the function under the contour integral of expression (16) will be analytic in $|z| < 1$. Hence, by virtue of the Cauchy's residue theorem this integral is zero as the contour of integration does not contain any singularity (Null integral property). Taking these considerations into account we can get a closed expression for $\sigma^2(z)$ as follows:

$$z(1-z)\frac{d^2\sigma^2(z)}{dz^2} - z\frac{d\sigma^2(z)}{dz} = 0 \quad \Rightarrow \tag{18}$$

$$\sigma^2(z) = 1 + z\left(1 + \log\left(\frac{z-1}{z}\right)\right)$$

### 3.3 Singularity Analysis

The basic principle of singularity analysis is the existence of a general correspondence between the asymptotic expansion of a function near its dominant singularities and the asymptotic expansion of the function's coefficients. Specifically, the method is mainly based on Cauchy's coefficient formula used in conjunction with special contours of integration known as Hankel contours [5]. Here we are interested in obtaining an asymptotic expression for the coefficients of the generating function obtained in (18). Hence, the first step consist in expressing such coefficients as a contour integral using the Cauchy's coefficient formula:

$$[z^n]\sigma^2(z) = \frac{1}{2\pi i}\oint_c \frac{\sigma^2(z)}{z^{n+1}}dz = \tag{19}$$

$$\frac{1}{2\pi i}\oint_c \frac{1}{z^{n+1}}dz + \frac{1}{2\pi i}\oint_c \frac{1}{z^n}dz + \frac{1}{2\pi i}\oint_c \frac{1}{z^n}\log\left(\frac{z-1}{z}\right)dz$$

The second step is to express the contour integral (19) using a Hankel contour. To this end, under the change of variables $z = 1 + t/n$, the kernel $z^{-n-1}$ in the integral (19) transform asymptotically into an exponential. Using the aforementioned change of variables in expression (19) together with a Hankel contour we obtain:

$$[z^n]\sigma^2(z) = \frac{1}{2\pi i n}\int_{+\infty}^{(0)}\left(1+\frac{t}{n}\right)^{-n-1}dt + \frac{1}{2\pi i n}\int_{+\infty}^{(0)}\left(1+\frac{t}{n}\right)^{-n}dt + \qquad (20)$$

$$+ \frac{1}{2\pi i n}\int_{+\infty}^{(0)}\left(1+\frac{t}{n}\right)^{-n}\log\left(\frac{t/n}{1+t/n}\right)dt$$

The contour and the associated rescaling capture the behaviour of the function near its singularities, enabling coefficient estimation when $n \to \infty$.

$$\frac{1}{\Gamma(s)} = \frac{-1}{2\pi i}\int_{+\infty}^{(0)}(-t)^s e^{-t}dt \qquad (21)$$

Re-arranging terms and expressing the integrals in terms of the gamma function (see expression 21), we finally get the asymptotic expansion for the variance of the error gradients:

$$\sigma_n^2 = [z^n]\sigma^2(z) \cong \frac{\gamma}{n} - \frac{1}{n^2} - \frac{1}{2n^3} \cong \frac{\gamma}{n} \qquad (22)$$

The result achieved by expression (22), where γ is the Euler number, is particularly interesting. It implies that for large enough values of n (data set size) we do not need to perform the iterative or the direct calculation of the variance (see expression (12) ) as it can be approximated using the above expression that is dependent exclusively on the size of the data set (or the size of the batch for sequential learning).

## 4  Practical Results

Expression (22) provides the law of asymptotic growth of the coefficients associated to the generating function describing the variance of the averaged error gradients. Figure 2 shows the result of computing the variance of a random variable. That is it shows the result of averaging *n* uniformly distributed random variables in the interval [0,1] for values of n (i.e. size of the data set), ranging from 500 up to 1000,000 (see the solid line style with squares) against the asymptotic approximation obtained in section 3 (see the dashed line style using diamonds at the sampling points).

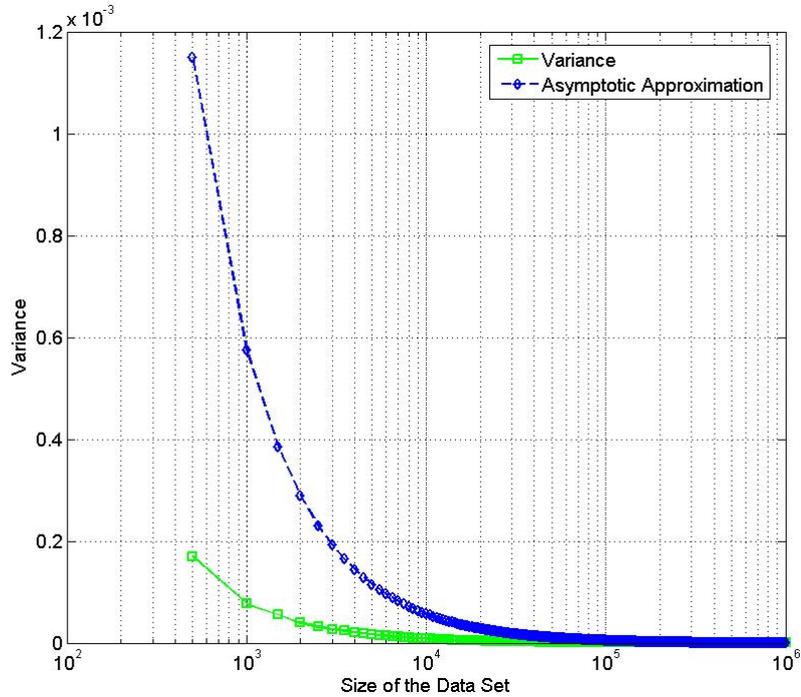

**Fig. 2.** The variance of the average of n (i.e. size of the data set) uniformly distributed random variables versus its asymptotic approximation…

From inspection of the graph, it can be observed that the approximation works quite well for values of *n* bigger or equal to 50,000. For example, at n = 49,500 the error between the real value and its approximation is less than $10^{-5}$. Similarly, for n = 5,000 the error is less than $10^{-4}$. These results are of particular interest for huge data sets (i.e. $n \geq 10^6$) where batch learning becomes impractical even for sophisticated gradient-based acceleration techniques like conjugate-gradients methods [2, 11] due to the memory storage requirements and associated computational cost. Furthermore, these kinds of acceleration techniques are not available for sequential learning. In this regard, vario-eta offers the possibility of working in sequential and batch learning modes.

Let us suppose that the size of the data set is *M*, and let us denote by *N* the size of the batch. The convergence of the learning rule in a sequential learning scenario is always guaranteed by the Robbins-Monro theorem [12] if the condition *M>>N* is satisfied. Therefore, under the hypothesis of a huge value of *M* if we further impose the condition *N>>1* the asymptotic approximation (22) will hold, thus we can obtain for a large-scale sequential learning problem a speed of convergence of an approximate second order algorithm at an extremely reduced computational cost. Nevertheless, it is important to note that these theoretical results must be interpreted carefully as real-world data sets usually contains correlated patterns that could break in certain cases the independence assumption made in section 3. Hence, providing a more practical guideline for real-world data sets remains for future work.

## 5 Conclusions

In this paper, we have investigated the applicability of an approximate second order stochastic learning rule to large-scale learning problems. Throughout this paper we have referred to the concept of structured representations of data as a way of increase the information content of a representation. In particular, within a machine learning context we have described the advantages of graph-based representation of images for classification purposes. However, this kind of representation requires new learning protocols able to deal not only with the increased complexity associated to the use of structured representations of information but also with huge learning data sets.

In this context, we have presented a mathematical description of the vario-eta learning rule. We have also assessed through a detailed analysis the reliability of its working hypothesis. Moreover, we have presented a careful complexity analysis of the algorithm aimed at understanding its asymptotic properties. As a result of this analysis we deduced an asymptotic expression for the learning rule. Specifically, such an approximation achieves a considerable reduction on the computational cost associated to the learning rule when dealing with large-scale learning problems.

**Acknowledgments:** The authors acknowledge the financial support by grant FIS 2009-9870 from the Spanish Ministry of Science and Innovation.

**References**

1. Bianchini, M., Gori, M, Sarti, L., Scarselli, F.: Recursive Processing of Cyclic Graphs. IEEE Transactions on Neural Networks 9 (17), pp. 10-18 (2006)
2. Bishop, C.M.: Neural Networks for Pattern Recognition, Oxford University Press, Oxford (1997)
3. Chinea, A.: Understanding the Principles of Recursive Neural Networks: A Generative Approach to Tackle Model Complexity. In: Alippi, C., Polycarpou, M., Panayiotou, C., Ellinas, G. (eds.) ICANN 2009. LNCS 5768, pp. 952-963. Springer, Heidelberg (2009)
4. Comtet, L.: Advanced Combinatorics: The Art of Finite and Infinite Expansions. Reidel Publishing Company, Dordrecht (1974)
5. Flajolet, P., Odlyzko, A. M.: Singularity Analysis of Generating Functions. In: SIAM Journal on Algebraic and Discrete Methods 3,2 New York (1990)
6. Flajolet, P., Sedgewick R.: Analytic Combinatorics. Cambridge University Press, Cambridge (2009)
7. Frasconi, P., Gori, M., Kuchler, A., Sperdutti, A.: A Field Guide to Dynamical Recurrent Networks. In: Kolen, J., Kremer, S. (Eds), pp. 351-364. IEEE Press, Inc., New York (2001).
8. Frasconi, P., Gori, M, Sperduti, A.: A General Framework for Adaptive Processing of Data Structures. IEEE Transactions on Neural Networks 9 (5), pp. 768-786 (1998)
9. Gallistel, C.R.: The Organization of Learning. MIT Press, Cambridge (1990)

10. Gori, M., Monfardini, G., Scarselli, L.: A New Model for Learning in Graph Domains. In: Proceedings of the 18[th] IEEE International Joint Conference on Neural Networks, pp. 729-734, Montreal (2005).
11. Haykin, S.: Neural Networks: a Comprehensive Foundation. Prentice Hall, New Jersey (1999)
12. Kushner, H.J., Yin, G. G.: Stochastic Approximation Algorithms and Applications. Springer-Verlag, New York (1997)
13. Leyton, M.: A Generative Theory of Shape. In: LNCS, vol 2145, pp. 1-76. Springer-Verlag (2001)
14. Leyton, M.: Symmetry, Causality, Mind. MIT Press, Massachusetts (1992).
15. Mahé, P., Vert, J.-P. : Graph Kernels Based on Tree Patterns for Molecules. In: Machine Learning, 75(1), pp. 3-35 (2009)
16. Mauro, C. D., Diligenti, M., Gori, M, Maggini, M.: Similarity Learning for Graph-based Image Representations. In: Pattern Recognition Letters, vol. 24, no. 8, pp. 115-1122, (2003)
17. Micheli, A., Sona, D., Sperduti, A.: Contextual Processing of Structured Data by Recursive Cascade Correlation. IEEE Transactions on Neural Networks 15(6) (2004) 1396-1410
18. Oppenheim, A. V., Schafer, R.: Discrete-Time Signal Processing. Prentice Hall, New Jersey (1989)
19. Tsochantaridis, I., Hofmann, T., Joachims, T., Altun, Y.: Support Vector Machine Learning for Interdependent and Structured Output Spaces. In: Brodley, C. E. (Ed.), ICML'04: Twenty-first international conference on Machine Learning. ACM Press, New York (2004)
20. Zimmermann, H. G., Neuneier, R.: How to Train Neural Networks. In: Orr, G.B., Müller, K.-R. (eds.): NIPS-WS 1996. LNCS, vol. 1524, pp. 395-399. Springer, Heidelberg (1998)